\documentclass[conference]{IEEEtran}
\IEEEoverridecommandlockouts
\usepackage{cite}
\usepackage{amsmath,amssymb,amsfonts}
\usepackage{algorithmic}
\usepackage{graphicx}
\usepackage{textcomp}
\usepackage{xcolor}
\usepackage{booktabs}
\usepackage{float}
\usepackage{multirow}
\usepackage{makecell}
\usepackage{hyperref}
\usepackage{adjustbox}
\usepackage{threeparttable}

\def\BibTeX{{\rm B\kern-.05em{\sc i\kern-.025em b}\kern-.08em
    T\kern-.1667em\lower.7ex\hbox{E}\kern-.125emX}}

\begin{document}

\title{SDA-GRIN for Adaptive Spatial-Temporal Multivariate Time Series Imputation\\
\thanks{
This work was undertaken thanks in part to funding from the Connected Minds program, supported by Canada First Research Excellence Fund, grant \#CFREF-2022-00010. This work was supported by NSERC Discovery RGPIN-2018-05550.}
}

\author{
\IEEEauthorblockN{1\textsuperscript{st} Amir Eskandari}
\IEEEauthorblockA{\textit{School of Computing} \\
\textit{Queen's University}\\
Kingston, ON, Canada \\
amir.eskandari@queensu.ca}
\and
\IEEEauthorblockN{2\textsuperscript{nd} Aman Anand}
\IEEEauthorblockA{\textit{School of Computing} \\
\textit{Queen's University}\\
Kingston, ON, Canada \\
aman.anand@queensu.ca}
\and
\IEEEauthorblockN{3\textsuperscript{rd} Drishti Sharma}
\IEEEauthorblockA{\textit{School of Computing} \\
\textit{Queen's University}\\
Kingston, ON, Canada \\
21ds128@queensu.ca}
\and
\IEEEauthorblockN{4\textsuperscript{th} Farhana Zulkernine}
\IEEEauthorblockA{\textit{School of Computing} \\
\textit{Queen's University}\\
Kingston, ON, Canada \\
farhana.zulkernine@queensu.ca}
}

\maketitle

\begin{abstract}
In various applications, the multivariate time series often suffers from missing data.\ This issue can significantly disrupt systems that rely on the data. Spatial and temporal dependencies can be leveraged to impute the missing samples.\ Existing imputation methods often ignore dynamic changes in spatial dependencies.\ We propose a Spatial Dynamic Aware Graph Recurrent Imputation Network (SDA-GRIN) which is capable of capturing dynamic changes in spatial dependencies.\ SDA-GRIN leverages a multi-head attention mechanism to adapt graph structures with time. SDA-GRIN models multivariate time series as a sequence of temporal graphs and uses a recurrent message-passing architecture for imputation.\ We evaluate SDA-GRIN on four real-world datasets: SDA-GRIN reduces MSE by 9.51\% for the AQI and 9.40\% for AQI-36.\ On the PEMS-BAY dataset, it achieves a 1.94\% reduction in MSE.\ Detailed ablation study demonstrates the effect of window sizes and missing data on the performance of the method. Project page: \url{https://ameskandari.github.io/sda-grin/}
\end{abstract}


\section{Introduction}\label{introduction}
Multivariate Time Series (MTS) data finds extensive applications in diverse domains, from healthcare \cite{lipton2016modeling} and geoscience \cite{percival2008analysis} to astronomy \cite{vaughan2013random}, and neuroscience \cite{gujral2020utilization}. MTS often exhibits numerous missing samples due to various factors, which can lead to deficiencies in systems that rely on this data. For example, malfunctioning sensors in traffic management systems can cause inaccurate data, leading to traffic congestion and negative environmental impact. In the Internet of Things (IoT), various interconnected devices generate data continuously, typically in the form of MTS, with each sensor representing a variable.  These sensors often exhibit missing values at any given time step. Since only a few sensors of all the deployed sensors on a system may have missing values, the dependencies can be used for imputation. 

Established methods for MTS imputation use both spatial and temporal dependencies for imputation \cite{marisca2022learning} \cite{you2020handling} \cite{cini2022filling}. These methods often use graphs to represent Spatial Dependencies (SD) among sensors. However, the construction of these graphs relies on assumptions about the nature of SD. These assumptions can be wrong due to the complex nature of MTS.\ Incorrect assumptions result from \textit{(i)} overlooking the dynamic nature of SD among variables and using a static graph to represent SD \cite{cini2022filling}; \textit{(ii)} employing similarity measurement functions that are not appropriate for the type of the data, e.g., using linear metrics for variables with non-linear relationships; \textit{(iii)} depending on geo-proximity information, that can be misleading, instead of relying on data-driven measurements \cite{aqi36-yi2016st}, e.g., two closely located sensors on opposite sides of a highway capture two different data streams on different traffic types.

To overcome these problems,\ we propose the Spatial Dynamic Aware Graph Recurrent Imputation Networks (SDA-GRIN).\ SDA-GRIN uses scaled dot-product Multi-Head Attention (MHA) \cite{vaswani2017attention} with a Message-Passing Recurrent Neural Network (MPRNN) \cite{cini2022filling} to effectively process and extract spatial and temporal dependencies for MTS imputation.\ The MHA attends to each variable and generates attention weights matrices, which are used to adapt the static adjacency matrix with time.\ The attention matrices capture the changes in SD through time. Next, the Message-Passing Gated Recurrent Network (MPGRU) \cite{li2017diffusion} module extracts spatial-temporal features. It uses a GRU-based architecture \cite{cho2014learning} where instead of a simple Multi-Layer Perceptron (MLP), a Message-Passing Neural Network (MPNN) is employed.\ We evaluate SDA-GRIN on four datasets: AQI and AQI-36 \cite{aqi36-yi2016st} \cite{Cao2018} from air quality domain, PEMS-BAY \cite{li2017diffusion} and METR-LA \cite{li2017diffusion} from traffic data domain.\ Our method outperforms previous state-of-the-arts \cite{cini2022filling} \cite{Cao2018} on all datasets.\ In summary, our contributions are as follows:\ (1) we propose SDA-GRIN,\ a spatial-temporal framework that enhances imputation by capturing dynamic SD changes,\ (2)\ we demonstrate significant performance improvements on METR-LA, PEMS-BAY, AQI, and AQI-36 benchmark datasets, and (3)\ the ablation study offers insights into the impact of missing data rate and window size on our propose models.

The rest of this paper is structured as follows: Section \ref{sec:RL} reviews related work on MTS imputation. Section \ref{sec:sdagrin} details the architecture and methodology of our proposed SDA-GRIN framework. Section \ref{experimental-design} presents the experimental setup, datasets, and analysis of the results, and Section \ref{conclusion} concludes the paper with a summary of findings.
\section{Related Work}\label{sec:RL}

MTS imputation has been widely studied in the literature.\ Traditional methods,\ like k-nearest neighbors \cite{beretta2016nearest},\ expectation-maximization\ \cite{kim2011incremental},\ support vector machines \cite{troyanskaya2001missing}, matrix factorization, and matrix completion \cite{yu2016temporal} \cite{mei2017nonnegative} techniques have been explored for this problem. State-space methods \cite{cai2022state} \cite{durbin2012time} focus on preserving the original structure of the data while filling in gaps.\ In recent years, deep learning techniques such as Generative Adversarial Networks (GANs) \cite{luo2018multivariate} and Recurrent Neural Networks (RNNs) \cite{che2018recurrent} \cite{Cao2018} have been used for MTS imputation. BRITS \cite{Cao2018} used a bidirectional recurrent-based architecture. In addition, transformer architecture has been used for this problem \cite{yildiz2022multivariate}.

\textbf{Graph-Based Methods:} Recent advances in MTS imputation and forecasting have leveraged GNNs \cite{cai2010graph} \cite{kipf2018neural} \cite{yu2017spatio} \cite{shang2021discrete} \cite{jin2024spatio} \cite{li2024dynamic} \cite{li2023dual} \cite{qian2024wavelet} \cite{eskandari2024gn2di}. Cini et al. \cite{cini2022filling} proposed GRIN, a two-step bidirectional MTS imputation framework. GRIN uses Message Passing Neural Networks (MPNN) instead of a simple MLP in GRU units \cite{cho2014learning}. GRIN demonstrates state-of-the-art performance for MTS imputation on various real-world benchmarks. However, it ignores changes in SD. Also,  Marisca et al. \cite{marisca2022learning} studied MTS imputation in sparse data. Chen et al. \cite{chen2022adaptive} developed AGRN as an approach for MTS forecasting. Similar to \cite{cini2022filling}, they replaced MLP in GRU units with their GCN-based architecture, enabling the method to extract both the spatial and temporal data features efficiently. Lan et al. \cite{lan2022dstagnn} developed DSTAGNN, which uses a stack of attention-based spatial-temporal blocks to extract features. Forecasting methods rely on one or multi-step-ahead auto-regressive prediction which does not produce accurate results for imputation tasks.

\section{SDA-GRIN: Overview}\label{sec:sdagrin}
We propose the Spatial Dynamic Aware Graph Recurrent Imputation Network (SDA-GRIN) for Multi-Variate Time Series Imputation (MTSI) built on top of GRIN \cite{cini2022filling}. SDA-GRIN models MTS as a sequence of temporal graphs, where each time step \( t \) is represented by a weighted directed graph \( \mathcal{G}_t \) with \( N \) nodes. Each graph \( \mathcal{G}_t = \langle\boldsymbol{X}_t, \boldsymbol{A}_t\rangle \) consists of a node-attribute matrix \( \boldsymbol{X}_t \in \mathbb{R}^{N \times T} \) and a weighted adjacency matrix \( \boldsymbol{A}_t \in \mathbb{R}^{N \times N} \).\ \( \boldsymbol{X}_t \) contains $T$ (window size) samples and $N$ variables, i.e., $x_{t,i}
 = \left[x_{(t-T+1),i},\cdots,x_{t,i}\right] \in R^{T}$, where $i\in \{1,\cdots,N\}$. SDA-GRIN processes \( \boldsymbol{X}_t \) and \( \boldsymbol{A}_t \) to extract relationships both within and across variables, utilizing a Message-Passing Neural Network (MPNN) with a recurrent processing mechanism.


Time order acts as a signal for a variable to extract and learn the temporal patterns.\ There is no explicit indicator that helps to understand the relationships among variables.\ Thus, we make assumptions and use graphs to represent the relationships among variables.\ These relationships can change over time and violate the assumptions.\ SDA-GRIN applies an adaptive mechanism to consider the changes in relationships using a Multi-Head Attention (MHA)-based module, as discussed in section \ref{subsec:multiheadatt}.\ We follow the GRIN \cite{cini2022filling} framework for the Spatial-Temporal (ST) feature extraction, processing, 
and imputation.\ Eq. \eqref{eq:s-enc} and Eq. \eqref{eq:mpgru} encode the spatial and ST features respectively using the GRU-based architecture by relying on the message-passing layers instead of MLPs. Eq. \eqref{eq:first-dec} and Eq. \eqref{eq:second-dec} perform the first and second imputation stages respectively. Eq. \eqref{eq:bidirect-imp} is used in bidirectional settings. 
\begin{equation}\label{eq:first-dec}
\widehat{X}_t^{(1)} = \Phi\left(\widehat{Y}_t^{(1)}\right), \, \widehat{Y}_{t}^{(1)} = H_{t-1}V_{h} + b_{h}
\end{equation}
\begin{equation}\label{eq:s-enc}
    S_{t} = \operatorname{MPNN}\left(\widehat{X}_t^{(1)},M_t,H_{t-1},A^{*}_{t} \right)
\end{equation}
\begin{equation}\label{eq:second-dec}
\widehat{X}_t^{(2)} = \Phi\left(\widehat{Y}_t^{(2)}\right), \, \widehat{Y}_t^{(2)} = \left[S_t \| H_{t-1}\right] V_s + b_s 
\end{equation}
\begin{equation}\label{eq:mpgru}
    H_{t} = \operatorname{MPGRU}\left(\widehat{X}^{(2)}_{t},M_t,H_{t-1},A^{*}_{t} \right)
\end{equation}
\begin{equation}\label{eq:bidirect-imp}
\widehat{y}^{i}_{t}=\operatorname{MLP}\left(\left[s^{i,fwd}_t\left\|h_{t-1}^{i,f w d}\right\| s_t^{i,b w d} \| h_{t+1}^{i,b w d}\right]\right)
\end{equation}
In \eqref{eq:first-dec} and \eqref{eq:second-dec}, $\Phi$ is the filtering operator, $\Phi(\boldsymbol{Y}_t) = \boldsymbol{M}_t \odot \boldsymbol{X}_t + \overline{\boldsymbol{M}}_t \odot \boldsymbol{Y}_t$.\ In the above equations, $\boldsymbol{M}_t \in R^{N\times T}$ represents the binary masking matrix, with $0$ for missing and $1$ for available samples. $V_h$, $b_h$, $V_s$, and $b_s$ are learnable parameters. $\widehat{X}^{(2)}_{t}$ and $\widehat{y}^{i}_{t}$ are the final results in the unidirectional, and the bidirectional settings, respectively. We use the bidirectional setting in SDA-GRIN. We train the model with mean absolute error (MAE) between the ground truth and imputed samples, calculating the loss across various imputation stages and directions (forward and backward).

\subsection{Spatial Dependency Awareness}\label{subsec:multiheadatt}
\begin{figure}
    \centering
    \includegraphics[width=1\linewidth]{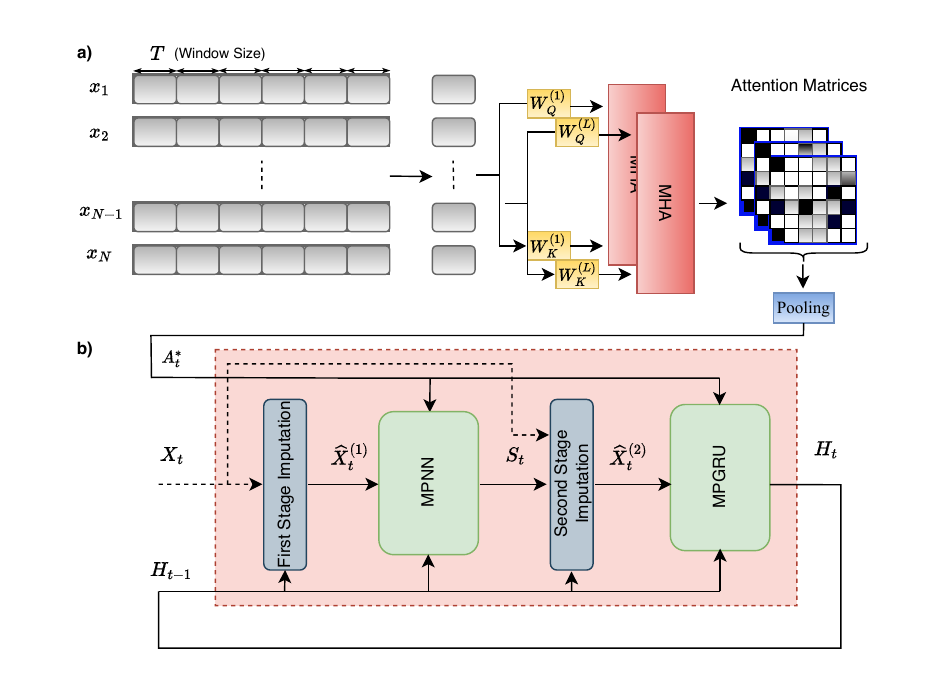}
    \caption{\small{Overview of SDA-GRIN. The multivariate time series (MTS) is chunked into windows. (a) MHA calculates attention among variables within each window, generating attention weights. These are pooled and used to adapt the static graph with time, creating $A^*_t$. (b) Later, in the unidirectional setting, the model processes the adapted graph ($A^*_t$), masked input ($X_t$), and previous context ($H_{t-1}$), outputting first-stage, second-stage imputed samples ($\hat{X}_t^{(1)}$, $\hat{X}_t^{(2)}$), and updated context ($H_t$).}}
    \label{fig:framework}
\end{figure}

We use MHA across variables over data samples within a fixed time window, $X_{t} \in \mathbb{R}^{N\times T}$ which has missing samples as zero, indicated by $M_t \in R^{N\times T}$. MHA extracts dynamic relationships among variables. For each variable, we can rewrite it as $\mathbf{x}_{t,i} \in \mathbb{R}^{T}$ where $i \in \{1,2,\ldots, N\}$. We calculate the attention weights between each pair of variables. First, we generate the query and key matrices as follows:
\begin{equation}
Q^{(l)}_{t} = X_{t}W^{(l)}_Q, \quad K^{(l)}_{t} = X_{t}W^{(l)}_K
\end{equation}
where $W^{(l)}_Q \in \mathbb{R}^{T\times d}$ and $W^{(l)}_K \in \mathbb{R}^{T\times d}$ are learnable matrices for the query and key, respectively, where $l$ denotes a head index. Next, we compute the attention weights:
\begin{equation}
\hat{A}^{(l)}_{t} = \text{softmax}\left(\frac{Q^{(l)}_{t}K^{(l)T}_{t}}{\sqrt{d}}\right)
\end{equation}
where $ \hat{A}^{(l)}_{t} \in \mathbb{R}^{N \times N} $ is the attention weights matrix for $l$-th head. In a multi-head setting, we have different $ \hat{A}^{(l)}_{t} $ for each head. We apply average pooling over the results of different heads.
\begin{equation}
\hat{A}_{t}^{\text{pooled}} = \frac{1}{L} \sum_{l=1}^{L} \hat{A}_{t}^{(l)}
\end{equation}
$L$ is the number of attention heads. MHA results in a dense matrix, making the graph fully connected, which can cause issues \cite{cini2022filling}.  In GNN layers, a sparse adjacency matrix is preferred. We use the static adjacency matrix in the datasets to make $\hat{A}_{t}^{\text{pooled}}$ sparse, $A_{t}^{*} = \hat{A}^{\text{pooled}}_{t} \odot (A > 0)$. $\odot$ denotes the element-wise multiplication and $A \in \mathbb{R}^{N \times N}$ is the static adjacency matrix which we calculate following previous methods \cite{Cao2018} \cite{aqi36-yi2016st} \cite{wu2019graph} \cite{li2017diffusion}. Fig. \ref{fig:framework} presents the overall framework.

\section{Experimental Design}\label{experimental-design}
\subsection{Dataset} We use four datasets from the Air Quality (AQ) and Traffic domains.\ The \textbf{AQI} dataset includes variables from 437 monitoring stations across 43 cities in China, representing different air quality indices. We use only the PM2.5 pollutant.\ Hourly measurements were collected from May 1, 2014, to April 30, 2015.\ The \textbf{AQI-36} dataset is a smaller version of the AQI dataset, containing only 36 sensors.\ We use the location information to generate the static adjacency matrix.\ We follow the same preprocessing, and evaluation settings from previous works \cite{Cao2018} \cite{aqi36-yi2016st}.\ Besides the AQ datasets, we also use the \textbf{PEMS-BAY} traffic dataset, which includes variables from 325 traffic sensors in San Francisco, and the \textbf{METR-LA} traffic dataset, containing 207 sensors from Los Angeles County highways.\ The sampling rate for both traffic datasets is 5 minutes.\ For traffic datasets, we randomly mask 25\% of the samples in each window and compute the static adjacency matrix following previous works \cite{wu2019graph} \cite{li2017diffusion}. 

\subsection{Implementation Details} We use the Adam \cite{kingma2014adam} optimizer with $1 \times 10^{-3}$ learning rate. A single NVIDIA A100 GPU is used with a batch size of 64 for AQI-36 and 256 for the other three datasets. We use a window size ($T$) of 256 for traffic datasets, 128 for AQI, and 32 for AQI-36. MHA uses 2 heads with 256 hidden dimensions for AQI, PEMS-BAY, and METR-LA and one head with 64 hidden dimensions for AQI36. To evaluate the performance of the proposed method, we compare its performance against MEAN, KNN \cite{beretta2016nearest}, MICE \cite{white2011multiple}, Matrix Factorization (MF), Vector Auto-regressive one-step-ahead predictor (VAR), BRITS \cite{Cao2018}, rGAIN \cite{miao2021generative}, MPGRU \cite{li2017diffusion}, and GRIN \cite{cini2022filling} as baselines.  

\begin{table}[b]
\centering
\scriptsize
\caption{\textbf{main results on AQ datasets}. The results for SDA-GRIN represent the mean of 5 runs. The baseline results are from \cite{cini2022filling}. We follow the same evaluation protocol from \cite{Cao2018} \cite{aqi36-yi2016st}. Underlined results indicate the second-best performance, while bolded results highlight the best-reported performance.}
\label{tab:result_air}
\resizebox{\columnwidth}{!}{%
\begin{threeparttable}
\begin{tabular}{cccccc}
\toprule
\textbf{Dataset} & \textbf{Method} & \textbf{MAE $\downarrow$} & \textbf{MSE $\downarrow$} & \textbf{MRE (\%) $\downarrow$} \\
\midrule
\multirow{7}{*}{AQI} 
 & Mean & 39.60 ± \footnotesize{0.00} & 3231.04 ± \footnotesize{0.00} & 59.25 ± \footnotesize{0.00} \\
 & KNN & 34.10 ± \footnotesize{0.00} & 3471.14 ± \footnotesize{0.00} & 51.02 ± \footnotesize{0.00} \\
 & VAR & 22.95 ± \footnotesize{0.30} & 1402.84 ± \footnotesize{52.63} & 33.99 ± \footnotesize{0.44} \\
 & BRITS & 20.21 ± \footnotesize{0.22} & 1157.89 ± \footnotesize{25.66} & 29.94 ± \footnotesize{0.33} \\
 & rGAIN & 21.78 ± \footnotesize{0.50} & 1274.93 ± \footnotesize{60.28} & 32.26 ± \footnotesize{0.75} \\
 & MPGRU & 18.76 ± \footnotesize{0.11} & 1194.35 ± \footnotesize{15.23} & 27.79 ± \footnotesize{0.16} \\
 \cmidrule(lr){2-5}
 & GRIN & \underline{14.73} ± \footnotesize{0.15} & \underline{775.91} ± \footnotesize{28.49} & \underline{21.82} ± \footnotesize{0.23} \\
 & \textbf{SDA-GRIN} & \textbf{14.43} ± \footnotesize{0.28} & \textbf{702.12} ± \footnotesize{22.82} & \textbf{21.59} ± \footnotesize{0.42} \\
\midrule
\multirow{7}{*}{AQI-36} 
  & Mean & 53.48 ± \footnotesize{0.00} & 4578.08 ± \footnotesize{0.00} & 76.77 ± \footnotesize{0.00} \\
 & KNN & 30.21 ± \footnotesize{0.00} & 2892.31 ± \footnotesize{0.00} & 43.36 ± \footnotesize{0.00} \\
 & BRITS & 14.50 ± \footnotesize{0.35} & 662.36 ± \footnotesize{65.16} & 20.41 ± \footnotesize{0.50} \\
 & rGAIN & 15.37 ± \footnotesize{0.26} & 641.92 ± \footnotesize{33.89} & 21.63 ± \footnotesize{0.36} \\
 & VAR & 15.64 ± \footnotesize{0.08} & 833.46 ± \footnotesize{13.85} & 22.02 ± \footnotesize{0.11} \\
 & MPGRU & 16.79 ± \footnotesize{0.52} & 1103.04 ± \footnotesize{106.83} & 23.63 ± \footnotesize{0.73} \\
\cmidrule(lr){2-5}
 & GRIN & \underline{12.08} ± \footnotesize{0.47} & \underline{523.14} ± \footnotesize{57.17} & \textbf{17.00} ± \footnotesize{0.67} \\
& \textbf{SDA-GRIN} & \textbf{12.05} ± \footnotesize{0.33} & \textbf{473.94} ± \footnotesize{34.65} & \underline{17.31} ± \footnotesize{0.47} \\
\bottomrule 
\end{tabular}
\end{threeparttable}}
\end{table}

\subsection{Results and Discussion}\label{experimental-results}
Tables \ref{tab:result_air}, and \ref{tab:result_traffic} show the performance of all the methods. We perform the training and testing for SDA-GRIN five times and report the mean and standard deviations and include the results of baselines from \cite{cini2022filling}. For the AQ datasets, as shown in Table \ref{tab:result_air}, SDA-GRIN demonstrates significant improvement. For AQI-36, our method achieves a 0.25\% reduction in Mean Absolute Error (MAE) and a 9.40\% reduction in Mean Squared Error (MSE). For AQI, SDA-GRIN reduces MAE by 2.04\%, MSE by 9.51\%, and Mean Relative Error (MRE) by 1.05\%. For the PEMS-BAY dataset (see Table \ref{tab:result_traffic}) our method achieves a reduction of 1.49\%, 1.94\%, and 0.93\%, on MAE, MSE, and MRE respectively compared to the best model in the literature \cite{cini2022filling}.\ Based on our observations, we believe that these reductions are mainly due to awareness of SD changes gained through MHA. Furthermore, SDA-GRIN's effectiveness across diverse datasets from multiple domains highlights its generalizability.

\begin{table}[t]
\centering
\scriptsize
\scriptsize 
\caption{\textbf{main results on traffic datasets.} The results for SDA-GRIN represent the mean of 5 runs. The baseline results are from \cite{cini2022filling}. The missing rate is 25\% for each variable. Underlined results show the second-best, while bolded indicate the best performance.}
\label{tab:result_traffic}
\begin{threeparttable}
\begin{adjustbox}{max width=\textwidth}
\begin{tabular}{cccccc}
\toprule
\textbf{Dataset} & \textbf{Method} & \textbf{MAE $\downarrow$} & \textbf{MSE $\downarrow$} & \textbf{MRE (\%) $\downarrow$} \\
\midrule
\multirow{10}{*}{PEMS-BAY} 
 & Mean & 5.42 ± \footnotesize{0.00} & 86.59 ± \footnotesize{0.00} & 8.67 ± \footnotesize{0.00} \\
 & KNN & 4.30 ± \footnotesize{0.00} & 49.80 ± \footnotesize{0.00} & 6.88 ± \footnotesize{0.00} \\
 & MF & 3.29 ± \footnotesize{0.01} & 51.39 ± \footnotesize{0.64} & 5.27 ± \footnotesize{0.02} \\
 & MICE & 3.09 ± \footnotesize{0.02} & 31.43 ± \footnotesize{0.41} & 4.95 ± \footnotesize{0.02} \\
 & VAR & 1.30 ± \footnotesize{0.00} & 6.52 ± \footnotesize{0.01} & 2.07 ± \footnotesize{0.01} \\
 & rGAIN & 1.88 ± \footnotesize{0.02} & 10.37 ± \footnotesize{0.20} & 3.01 ± \footnotesize{0.04} \\
 & BRITS & 1.47 ± \footnotesize{0.00} & 7.94 ± \footnotesize{0.03} & 2.36 ± \footnotesize{0.00} \\
 & MPGRU & 1.11 ± \footnotesize{0.00} & 7.59 ± \footnotesize{0.02} & 1.77 ± \footnotesize{0.00} \\
 \cmidrule(lr){2-5}
 & GRIN & \underline{0.67} ± \footnotesize{0.00} & \underline{1.55} ± \footnotesize{0.01} & \underline{1.08} ± \footnotesize{0.00} \\
 & \textbf{SDA-GRIN} & \textbf{0.66} ± \footnotesize{0.00} & \textbf{1.52} ± \footnotesize{0.01} & \textbf{1.07} ± \footnotesize{0.00} \\
\midrule
\multirow{10}{*}{METR-LA} 
 & Mean & 7.56 ± \footnotesize{0.00} & 142.22 ± \footnotesize{0.00} & 13.10 ± \footnotesize{0.00} \\
 & KNN & 7.88 ± \footnotesize{0.00} & 129.29 ± \footnotesize{0.00} & 13.65 ± \footnotesize{0.00} \\
 & MF & 5.56 ± \footnotesize{0.03} & 113.46 ± \footnotesize{1.08} & 9.62 ± \footnotesize{0.05} \\
 & VAR & 2.69 ± \footnotesize{0.00} & 21.10 ± \footnotesize{0.02} & 4.66 ± \footnotesize{0.00} \\
 & MICE & 4.42 ± \footnotesize{0.07} & 55.07 ± \footnotesize{1.46} & 7.65 ± \footnotesize{0.12} \\
 & BRITS & 2.34 ± \footnotesize{0.00} & 16.46 ± \footnotesize{0.05} & 4.05 ± \footnotesize{0.00} \\
 & rGAIN & 2.83 ± \footnotesize{0.01} & 20.03 ± \footnotesize{0.09} & 4.91 ± \footnotesize{0.01} \\
 & MPGRU & 2.44 ± \footnotesize{0.00} & 22.17 ± \footnotesize{0.03} & 4.22 ± \footnotesize{0.00} \\
 \cmidrule(lr){2-5}
 & GRIN & \textbf{1.91} ± \footnotesize{0.00} & \textbf{10.41} ± \footnotesize{0.03} & \textbf{3.30} ± \footnotesize{0.00} \\
 & \textbf{SDA-GRIN} & \textbf{1.91} ± \footnotesize{0.00} & \underline{10.48} ± \footnotesize{0.04} & \textbf{3.30} ± \footnotesize{0.01} \\
\bottomrule
\end{tabular}
\end{adjustbox}
\end{threeparttable}
\end{table}

\begin{figure}[t]
    \centering
    \includegraphics[width=1\linewidth]{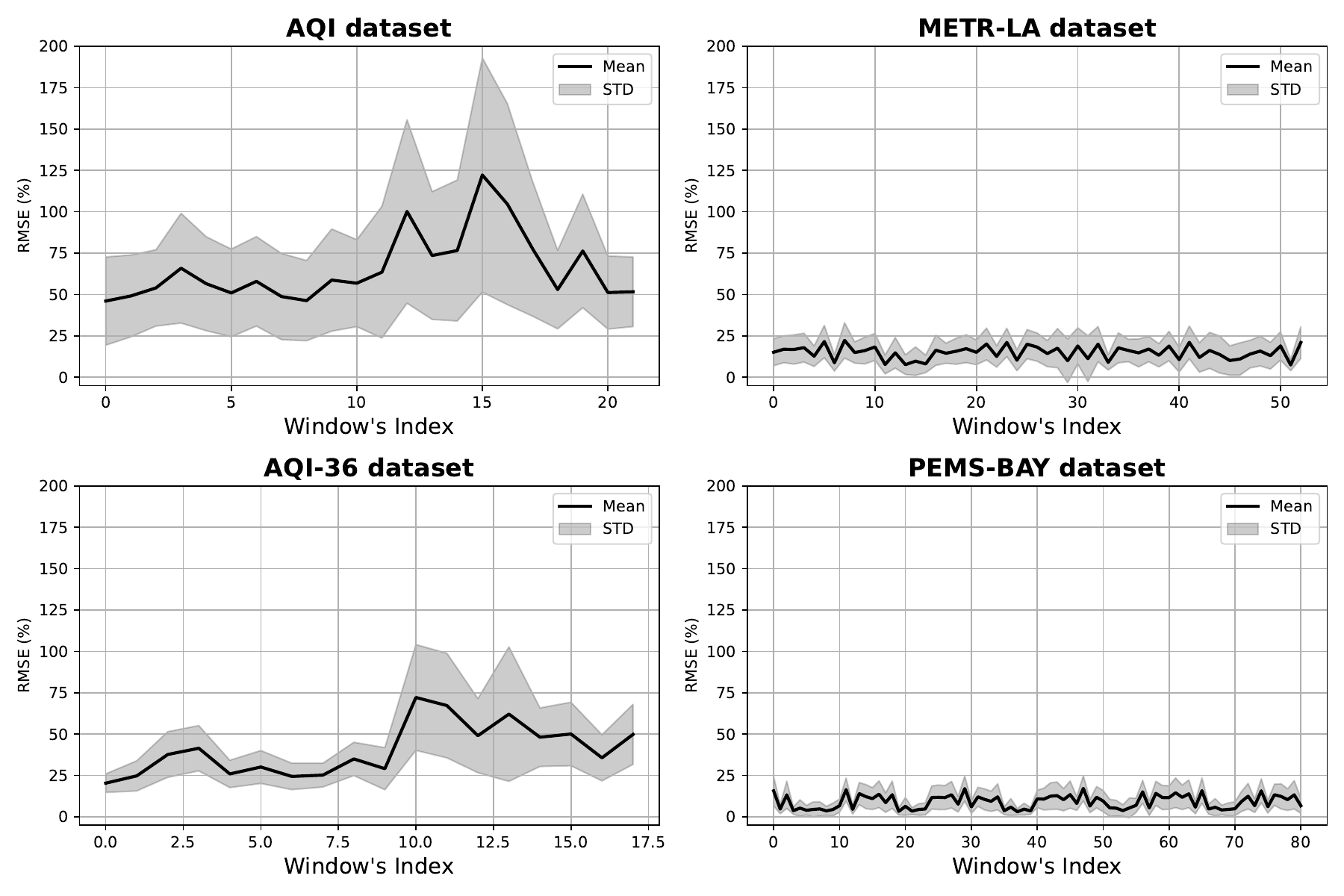}
    \caption{Pair-wise Relative Mean Squared Error (RMSE) among all variables at each step, shown for AQI, AQI-36, PEMS-BAY, and METR-LA datasets. We use 128 as the window size for the validation part of the datasets. The line represents the mean, while the margin indicates the standard deviation of pair-wise RMSE of all variables. AQI and AQI-36 datasets show greater values and variation within a window and across different windows compared to METR-LA and PEMS-BAY.}
    \label{fig:rmse-datasets}
\end{figure}

Results for AQ datasets are shown in Table \ref{tab:result_air}.\ As we can see, 
SDA-GRIN achieves better results than traffic datasets.\ This is mainly due to the different nature of datasets as shown in Fig. \ref{fig:rmse-datasets} which shows the mean and standard deviation of pair-wise Relative Mean Squared Error (RMSE) among all variables for different window indices. We use the validation part of the datasets for Fig. \ref{fig:rmse-datasets}. It shows that pair-wise distances among variables in AQ datasets are larger and they change significantly at different time steps. The large variation in the plot of the AQI and AQI-36 dataset indicates that spatial relations are volatile and change frequently which results in greater RMSE compared to the traffic datasets.\ Therefore, adaptation is crucial for such data.\ SDA-GRIN effectively addresses this while maintaining minimal computational overhead (see Appendix B for details).

\begin{figure}[t]
    \centering
    \includegraphics[width=1\linewidth]{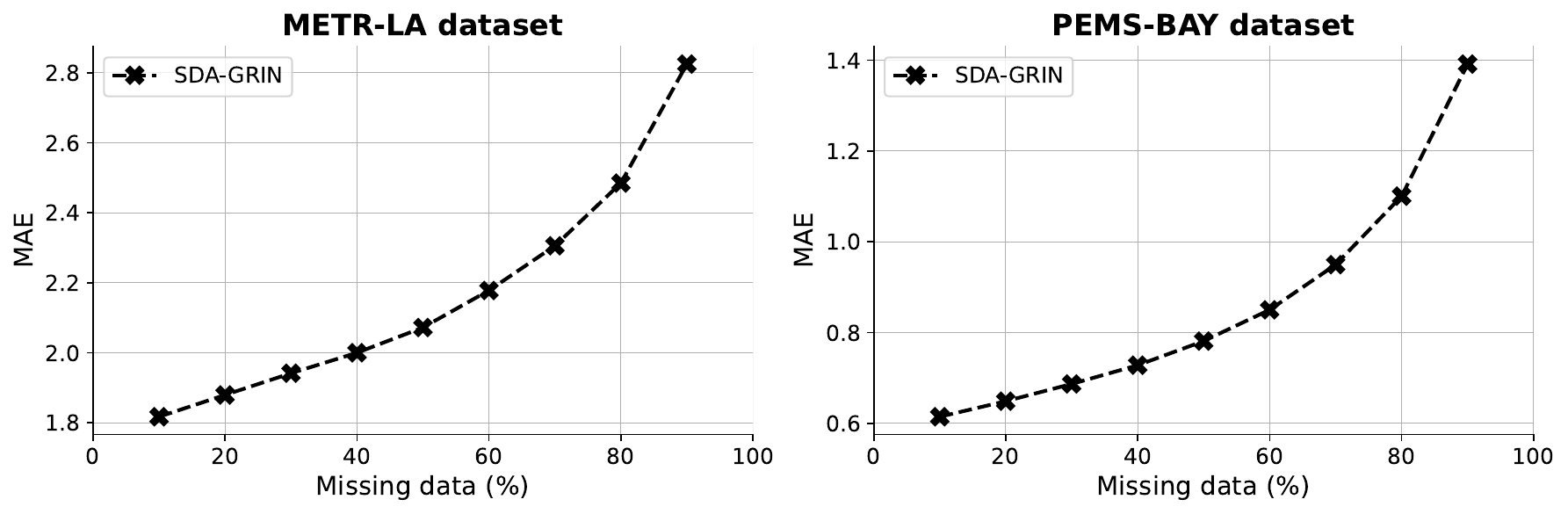}
    \caption{Performance of SDA-GRIN across different missing rates, ranging from 10\% to 90\% in 10\% increments. Performance declines at higher missing rates due to the reduced availability of samples within each variable for MHA to attend to.}
    \label{fig:missing_rate}
\end{figure}

\textbf{Ablation Study.}
The two key parameters in the experiments that affect the method's performance are window size (see Table \ref{tab:ablation_study}) and missing data rate (see Fig. \ref{fig:missing_rate}). The window size determines the number of samples of variables that the MHA can attend to at each step. We experiment with four different window sizes: $T=\{32,64,128,254\}$.\ As shown in Table \ref{tab:ablation_study}, For PEMS-BAY, METR-LA, and AQI, the largest window size performs the best. We attribute this to the fact that a larger context length allows MHA to better understand the dynamics.\ However, this is not the case for the AQI-36 dataset, where the smallest window size shows the best performance. We believe that this observation is due to the dataset's low number of variables (36) compared to greater than 200 variables in the other datasets. For the missing rate, we experimented SDA-GRIN with missing rate values ranging from $10\%$ to $90\%$ with PEMS-BAY and METR-LA datasets. Fig. \ref{fig:missing_rate} shows that SDA-GRIN's performance declines as the missing rate increases. This decline occurs because, at higher missing rates, most samples of variables are filled with zeros (representing missing values), making it difficult for the MHA mechanism to detect changes and adapt the graph structure effectively. See Appendix A for the ablation study on MHA attention heads.

\begin{table}[t]
\centering
\scriptsize 
\caption{\textbf{Effect of window size.} From the perspective of MHA, window size refers to the context length used for attending to and adapting the graph structure. The bolded results indicate the best performance. $^{*}$ For the AQI dataset, the window size of 256 caused the GPU to overflow due to the high number of variables (437). }
\label{tab:ablation_study}
\begin{threeparttable}
\begin{adjustbox}{max width=\columnwidth}
\begin{tabular}{cccccc}
\toprule
\textbf{Dataset} & \textbf{Window Size} & \textbf{MAE $\downarrow$} & \textbf{MSE $\downarrow$} & \textbf{MRE (\%) $\downarrow$} \\
\midrule
\multirow{4}{*}{AQI} & 32 & 14.35 ± \footnotesize{0.09} & 701.19 ± \footnotesize{6.74} & 21.48 ± \footnotesize{0.14} \\
 & 64 & 14.44 ± \footnotesize{0.28} & 701.08 ± \footnotesize{22.82} & 21.60 ± \footnotesize{0.42} \\
 & 128 & \textbf{14.29 ± \footnotesize{0.07}} & \textbf{684.83 ± \footnotesize{5.93}} & \textbf{21.38 ± \footnotesize{0.11}} \\
 & 256$^{*}$ & --- & --- & --- \\
\midrule
\multirow{4}{*}{AQI-36} & 32 & \textbf{12.06 ± \footnotesize{0.33}} & 473.94 ± \footnotesize{34.65} & \textbf{17.31 ± \footnotesize{0.47}} \\
 & 64 & 12.06 ± \footnotesize{0.20} & \textbf{460.42 ± \footnotesize{9.17}} & 17.32 ± \footnotesize{0.29} \\
 & 128 & 12.66 ± \footnotesize{0.25} & 476.53 ± \footnotesize{30.19} & 18.17 ± \footnotesize{0.36} \\
 & 256 & 13.27 ± \footnotesize{0.30} & 520.02 ± \footnotesize{35.68} & 19.05 ± \footnotesize{0.42} \\
\midrule
\multirow{4}{*}{METR-LA} & 32 & 1.97 ± \footnotesize{0.00} & 10.98 ± \footnotesize{0.01} & 3.42 ± \footnotesize{0.00} \\
 & 64 & 1.96 ± \footnotesize{0.00} & 10.86 ± \footnotesize{0.03} & 3.39 ± \footnotesize{0.00} \\
 & 128 & 1.93 ± \footnotesize{0.00} & 10.60 ± \footnotesize{0.03} & 3.35 ± \footnotesize{0.00} \\
 & 256 & \textbf{1.91 ± \footnotesize{0.00}} & \textbf{10.49 ± \footnotesize{0.04}} & \textbf{3.30 ± \footnotesize{0.01}} \\
\midrule
\multirow{4}{*}{PEMS-BAY} & 32 & 0.68 ± \footnotesize{0.00} & 1.54 ± \footnotesize{0.01} & 1.09 ± \footnotesize{0.00} \\
 & 64 & 0.68 ± \footnotesize{0.00} & 1.52 ± \footnotesize{0.01} & 1.08 ± \footnotesize{0.00} \\
 & 128 & 0.67 ± \footnotesize{0.00} & \textbf{1.51 ± \footnotesize{0.01}} & 1.07 ± \footnotesize{0.00} \\
 & 256 & \textbf{0.67 ± \footnotesize{0.00}} & 1.53 ± \footnotesize{0.01} & \textbf{1.07 ± \footnotesize{0.00}} \\
\bottomrule
\end{tabular}
\end{adjustbox}
\end{threeparttable}
\end{table}


\section{Conclusion}\label{conclusion}
In this paper, we propose SDA-GRIN, to enhance spatial-temporal feature extraction to understand changes of spatial dependencies among multiple variables in time series data for missing value imputation. Using static spatial dependencies is not enough when relations change over time. We use a multi-head attention-based mechanism to adapt to changes in spatial dimension. The empirical results demonstrate that SDA-GRIN achieves significant improvement over previous baselines on datasets from traffic and air quality domains. We achieved 2.04\% MAE and 9.51\% MSE reduction on AQI, 0.25\% MAE and 9.40\% MSE reduction on AQI-36, and a 1.49\% MAE and 1.94\% MSE reduction on PEMS-BAY. SDA-GRIN particularly is more effective when the relationships among variables in the data have high variance over time. Our ablation study also demonstrates that greater window size can capture more context for adaptation mechanisms for datasets having a comparable number of variables. Also, it indicates that the adaptation mechanism is sensitive to high levels of missing data, as the MHA struggles to capture changes when only a few samples are available within a given window.

\bibliographystyle{IEEEtran}
\bibliography{refrences}

\appendices

\section{Ablation on Number of Attention Heads}
To determine the optimal number of attention heads ($L$) in the Multi-Head Attention (MHA) module, we conducted a random search over different values. Since increasing the number of heads adds computational overhead, we intentionally kept the search space limited to a small number of heads.
\begin{table}[h]
\centering
\small 
\caption{Ablation Study Results on Attention Heads}
\begin{tabular}{@{}lcccc@{}}
\toprule
Dataset & Heads & MAE $\downarrow$ & MSE $\downarrow$ & MAPE $\downarrow$ \\
\midrule
\multirow{4}{*}{AQI36} 
& 1 & \textbf{12.19} & \textbf{461.80} & 0.4247 \\
& 2 & 12.28 & 483.42 & \textbf{0.4097} \\
& 3 & 12.37 & 464.73 & 0.4340 \\
& 4 & 14.09 & 519.04 & 0.6130 \\
\midrule
\multirow{3}{*}{AQI}
& 1 & 14.77 & 748.67 & \textbf{0.3576} \\
& 2 & \textbf{14.41} & \textbf{708.47} & 0.3640 \\
& 4 & 14.78 & 721.88 & 0.3889 \\
\bottomrule
\end{tabular}
\label{tab:ablation_heads}
\end{table}

Table \ref{tab:ablation_heads} presents the results for AQI and AQI-36 datasets. Notably, for AQI-36, using a single attention head resulted in the lowest MAE and MSE, while two heads achieved the best MAPE. However, increasing the number of heads beyond two led to degraded performance across all metrics. A similar trend was observed for AQI, where two heads achieved the lowest MAE and MSE, while one head had the best MAPE. This suggests that a smaller number of attention heads is sufficient for capturing spatial dependencies in MTS imputation.

\section{Computational Overhead}
SDA-GRIN introduces minimal computational overhead, adding only the attention matrices \( W^{(l)}_{Q} \) and \( W^{(l)}_{K} \) (each of size \( T \times D \)) with just 1-2 attention heads per dataset to the backbone model, GRIN. As shown in Table \ref{tab:parameters}, this results in only a 1.1–4.9\% increase in total parameters across all datasets. These significant performance gains with minimal parameter growth highlight the efficiency of our approach.
\begin{table}[H]
\centering
\caption{Parameter comparison of GRIN and SDA-GRIN models}
\resizebox{\columnwidth}{!}{  
\begin{tabular}{lrrr}
\toprule
Dataset & Original (K) & Added (K) & Increase (\%) \\
\midrule
AQI & 607 & 16 & 2.7 \\
AQI-36 & 214 & 2 & 1.1 \\
METR-LA & 674 & 33 & 4.9 \\
PEMS-BAY & 690 & 33 & 4.8 \\
\bottomrule
\end{tabular}
}
\label{tab:parameters}
\end{table}

\section{Comparison Between Static and Pooled Graphs}
\begin{figure}[H]
    \centering
    \includegraphics[width=1.0\linewidth]{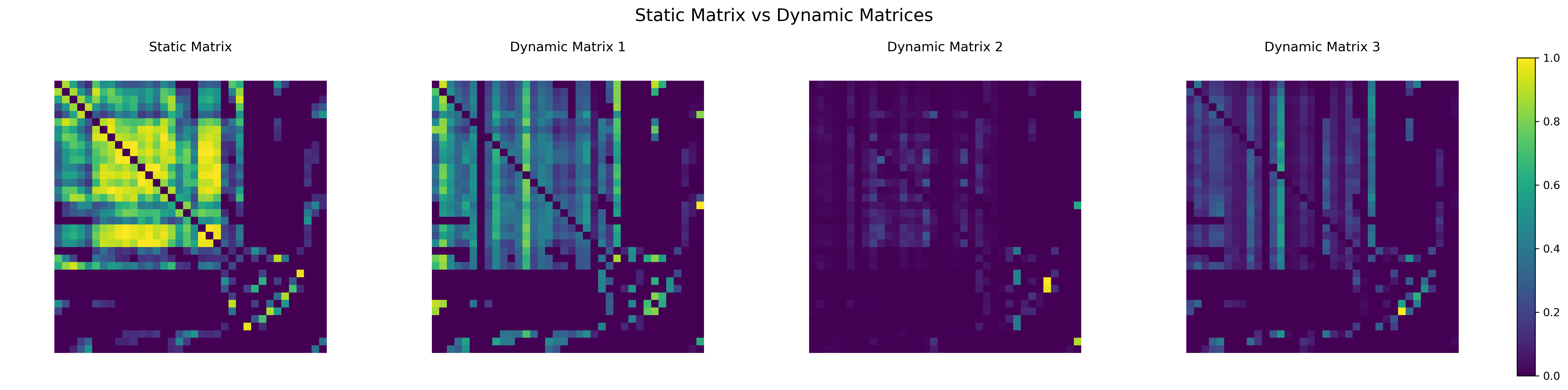}
    \caption{Comparison of the static adjacency graph with three random dynamic graphs for AQI36.}
    \label{fig:statvsdynamic}
\end{figure}

To investigate the benefits of SDA-GRIN’s adaptive graph mechanism, we compare a static adjacency matrix with three dynamic matrices from different time windows of the AQI-36 dataset, as illustrated in Fig. \ref{fig:statvsdynamic}. The static graph captures fixed spatial dependencies, but in multivariate time series data, such dependencies are rarely stationary. In contrast, the dynamic matrices generated through Multi-Head Attention (MHA) in SDA-GRIN show significant variations, revealing evolving relationships. These dynamic graphs effectively capture time-sensitive spatial dependencies, enhancing imputation performance in datasets with high spatial variability.

\end{document}